# XTDrone: A Customizable Multi-Rotor UAVs Simulation Platform

Kun Xiao[1], Shaochang Tan[2], Guohui Wang[1], Xueyan An[3], Xiang Wang[4], Xiangke Wang[5]
[1] Beijing Institute of Aerospace Systems Engineering, Beijing, China
[2] Flying College, Beihang University, Beijing, China
[3] China Academy of Launch Vehicle Technology, Beijing, China
[4] College of Architecture and Urban Planning, Tongji University, Shanghai, China
[5] College of Intelligence Science and Technology, National University of Defense Technology, Changsha, China
e-mail: xkwang@nudt.edu.cn

*Abstract*—A customizable multi-rotor UAVs simulation platform based on ROS, Gazebo and PX4 is presented. The platform, which is called XTDrone, integrates dynamic models, sensor models, control algorithm, state estimation algorithm, and 3D scenes. The platform supports multi UAVs and other robots. The platform is modular and each module can be modified, which means that users can test its own algorithm, such as SLAM, object detection, motion planning, attitude control, multi-UAV cooperation, and cooperation with other robots on the platform. The platform runs in lockstep, so the simulation speed can be adjusted according to the computer performance. In this paper, two cases, evaluating different visual SLAM algorithm and realizing UAV formation, are shown to demonstrate how the platform works. XTDrone is all open source [1] and a detailed manual is provided [2].

*Keywords-multirotor UAV;customizable;simulation platform; visual SLAM; UAV formation*

## I. INTRODUCTION

Multi-rotor UAVs (unmanned aerial vehicles) have been widely used in the civil and military market in recent years due to its outstanding maneuverability, stability and hover performance [1]. Nowadays, multi-rotor UAVs are expected to be more and more intelligent. In addition to advanced control and state estimation, computer vision and artificial intelligence are widely applied. Compared to cars and tradition robots, UAVs are more likely to be out of control. Debugging algorithm on real UAVs is time-consuming and dangerous. And for reinforcement learning, it's quite difficult to train UAVs directly in the physical world. Therefore, a general simulation platform plays an important role in the algorithm design and deployment.

There are some powerful flight simulators which support multi-rotor UAVs, such as X-Plane [2], FlightGear [3] and Airsim [4]. Plenty of researchers and engineers use these platform to study and design UAVs. Powerful physics engine and 3D graph engine make these simulators close to physical world. However, they are flight simulator, not robot simulator, which means that they are not proper for general robot mission. To some extent, UAVs are more like robots than aircraft. For example, multi-rotor UAVs equipped with robot arms are a hot research now [5], but it's hard to realize this simulation in flight simulators mentioned above.

The Robot Operating System (ROS) is a set of software libraries and tools to build robot applications and is all open source [6]. From drivers to state-of-the-art algorithms, and with powerful developer tools, ROS makes robot design convenient. And Gazebo [7], an open-source 3D robotics simulator, is the main simulator of ROS. ROS together with Gazebo, is very popular in the robotics area. In addition, there are some useful UAV models and libraries for Gazebo and ROS thanks to some researchers [8] [9]and commercial companies.

PX4 [9], one of the most popular open source UAV autopilot software, helps developers solve the embedding communication and control, so they can focus on high level function, such as perception and planning. PX4 can be deployed not only in embedding system, but also in computers, which is called Software In the Loop (SITL) simulation. PX4 SITL can communication with Gazebo by MAVLink [3] and communicate with ROS by MAVROS [4].

Based on ROS, Gazebo and PX4, a modular multi-rotor UAVs simulation platform called XTDrone was developed. Fig. 1 shows it's architecture. Modules in blue are Gazebo modules, modules in yellow are PX4 SITL (software in the loop) modules and modules in pale orange are other modules. Each module can be modified conveniently, so users can test its own algorithm on the platform. Because ROS and PX4 is originally designed for embedding system, users can deploy their algorithm to real UAVs conveniently after testing and debugging on the simulation platform. Moreover, the platform runs in lockstep, so the simulation speed can be adjusted according to the computer performance.

Although XTDrone supports control and state estimation design, this paper focuses on perception, motion planning and cooperation. For developing and testing control and state estimation algorithms, developers should learn more about PX4 [5].

The remainder of this paper is organized as follows: In Section II, UAV dynamic system is established. In Section III, sensor models are established. And then, three demos are presented to show how the platform works. In Section IV, we evaluate ORB-SLAM2 [10] and VINS-Fusion without IMU [11], comparing their localization accuracy and computational complexity. In Section V, we realize UAV formation and reconfiguration. In Section VI, conclusion is made and future work is discussed. Besides two demos presented in this paper, other demos such as laser SLAM, VIO, and object detection and tracking, can be seen at gitee[1].

---

[1] https://gitee.com/robin_shaun/XTDrone or
https://github.com/robin-shaun/XTDrone
[2] https://www.yuque.com/xtdrone/manual_en
[3] https://mavlink.io/en/
[4] http://wiki.ros.org/mavros
[5] https://dev.px4.io/master/en/index.html

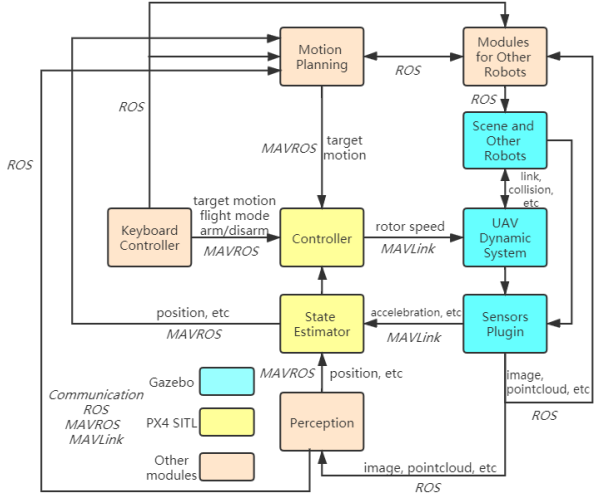

Figure 1. Architecture of XTDrone

## II. UAV DYNAMIC SYSTEM

For UAVs, dynamic system is highly relative to control system. Therefore, the accuracy of dynamic system is an important standard to evaluate a simulation platform. This section will introduce UAV dynamic system, including dynamic equations and the numerical solver.

In the simulation environment, UAV is idealized as a 6-DoF rigid body. Its sketch is shown in Fig. 2. $v$ is the flight speed. The forces and moments on an UAV are mainly composed of the gravity of the entire aircraft $G$, the aerodynamic forces $F_i$ and moments $M_i$ generated by each single rotor, and the air drag $D$ of the fuselage. Thus for a UAV with n rotors, Newton's law (1) and Euler's equation (2) on the Center of Gravity (CoG) can be written as:

$$\sum_{i=1}^{n}(F_i + D + G) = ma \quad (1)$$

$$\sum_{i=1}^{n}(M_i + r_i \times F_i) = J \cdot \dot{\Omega} + \Omega \times (J \cdot \Omega) \quad (2)$$

where m is the mass of the UAV, $a$ is its acceleration, $\Omega$ is its angular velocity, $J$ is its inertia matrix, and $r_i$ is a vector from the CoG to rotor i.

For many simulation experiments are now mainly performed at low speed and low wind, which means the drag on the fuselage $D$ can be neglected. However, developers can easily enable it by adding a wind plugin to their model file and set correct aerodynamic coefficients [6]. And in order to conveniently apply these forces and moments in the simulator the rotors' force $F_i$ are decomposed into the thrust force $T_i$ and the H-force $H_i$, where H-force is often used to describe drag perpendicular to thrust lying on the rotor plane in the helicopter literature [12]; $M_i$ are decomposed into the rolling moment $M_{R,i}$ and the moment due to drag of a rotor blade $M_{D,i}$ [13].

$$T_i = -\omega_i^2 C_T e_{z,b} \quad (3)$$

$$H_i = -\omega_i C_D v^{\perp} \quad (4)$$

$$M_{R,i} = -\zeta \omega_i C_R v^{\perp} \quad (5)$$

$$M_{D,i} = -\zeta C_M T_i \quad (6)$$

$\omega_i$ denotes the angular velocity of rotor i, $e_{z,b}$ is a unit vector that points in the z-direction in body frame. And $v^{\perp}$ denotes the projection of UAV's velocity onto the rotor plane. $\zeta$ denotes the turning direction, it can be +1 (CCW) or −1 (CW). $C_T$, $C_D$, $C_R$, $C_M$ are all positive constants. And following (7)-(10), developers can conveniently adjust those constants to suit their own UAV models [14].

$$C_T = \frac{C_{t0} \rho d^4}{(2\pi)^2} \quad (7)$$

$$C_D = \frac{1}{16} \rho m c C_{d0} d^2 \quad (8)$$

$$C_R = (\frac{\theta_0}{48} + \frac{\theta_1}{64}) \rho n k c d^3 \quad (9)$$

$$C_M = \frac{C_{m0}}{C_{t0}} d \quad (10)$$

Where $\rho$ is the air density, $k$ is the Lift-curve slope of the blade airfoil, $d$ is the rotor diameter, $m$ denotes the number of blades of the rotor, and $c$ is the average chord length. $C_{t0}$ is the static thrust coefficient, $C_{d0}$ is the static drag coefficient, and $C_{m0}$ is the static moment coefficient. $\theta_0$ denotes the pitch angle of blade root, and $\theta_1$ denotes the twist of the blade.

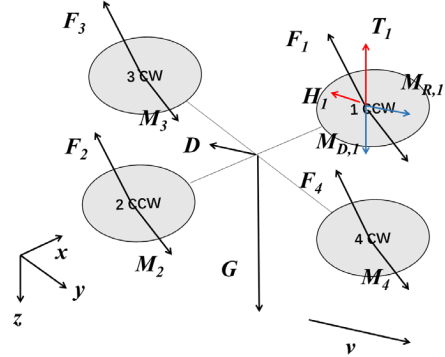

Figure 2. Sketch of the UAV

For numerical solver, XTDrone uses Open Dynamics Engine (ODE) [7]. ODE is an open source, high performance library for simulating rigid body dynamics. It is fully featured, stable, mature and platform independent with an easy to use C/C++ API. It has advanced joint types and integrated collision detection with friction.

---

[6] https://www.yuque.com/xtdrone/manual_en/wind_plugin

[7] https://www.ode.org/

## III. SENSOR MODELS

For UAVs, sensor models are highly relative to state estimation, so it's important to supply accurate sensor models for a simulation platform. This section will introduce models of IMU (inertial measurement unit), magnetometer, barometer, GPS (Global Positioning System), and stereo camera. More sensors such as lidar and depth camera can be seen at XTDrone manual [8].

### A. IMU, magnetometer, barometer and GPS

IMU, consisting of accelerometer and gyroscope, and measuring acceleration and angular velocity, is the smallest unit for UAV attitude estimation [15]. For PX4 state estimation algorithm, magnetometer to measure three-axis angle and barometer to measure height are added to improve performance.

State estimation with IMU, magnetometer and barometer can supply attitude and vertical position, but can't supply horizontal position. Therefore, GPS, which supplies global position and velocity, is quite popular for UAVs.

Although these sensors have different principle of measurement, they have a similar simplified measurement model.

$$x_m = x + b + n \qquad (11)$$

$$\dot{b} = n_b \qquad (12)$$

Where $x_m$ is the measured value (such as x-axis acceleration, y-axis angular velocity and z-axis position), $x$ is ground truth, $b$ is bias which random walks, $n$ and $n_b$ are Gaussian noise, $n$ is white noise but $n_b$ is not because the distribution of $n_b$ is time-correlation [16]. Users can set parameters of bias and white noise according to the real sensor datasheet. Parameters are listed in Table I.

TABLE I. PARAMETERS OF IMU, MAGNETOMETER, BAROMETER AND GPS

| Parameters | Description |
|---|---|
| NoiseDensity | Standard deviation of $n$ |
| RandomWalk | Standard deviation of $n_b$ |
| BiasCorrelationTime | Correlation time of $b$ |
| TurnOnBiasSigma | Initial standard deviation of $b$ |

### B. Stereo camera

Even though GPS can supply global position, but its precision is usually low. And most importantly, in certain environment, such as indoor environment, little GPS signal can be received. Vision-based navigation is an effective aid to GPS navigation. And stereo cameras are common sensors for vision-based navigation.

A stereo camera is a type of camera with two or more lenses with a separate image sensor [17]. For UAVs, commonly there are two lenses placed horizontally, and the distance between two lenses is called length of baseline. Except the length of baseline, there are many parameters for each single camera. The main parameters are listed in Table II. Ref. [18] introduces about the parameters in detail.

TABLE II. PARAMETERS OF STEREO CAMERA

| Parameters | | Description |
|---|---|---|
| image | Width, Height | Width, Height in pixels |
| | format | Format of RGB space |
| noise | mean | The mean of Gaussian noise |
| | stddev | The stand deviation of Gaussian noise |
| intrinsics | fx, fy | X, Y focal length |
| | cx, cy | X, Y principal point |
| | skew | XY axis skew |
| distortion | k1, k2, k3 | Radial distortion coefficient |
| | p1, p2 | Tangential distortion coefficient |
| | center | Distortion center or principal point |
| clip | Near, far clipping | Near and far clip planes |
| hackBaseline | | Length of baseline |

## IV. VISUAL SLAM EVALUATION

Simultaneous localization and mapping (SLAM) is the computational problem of constructing or updating a map of an unknown environment while simultaneously keeping track of an agent's location within it [19]. SLAM includes visual SLAM and laser SLAM, and in this section, two visual SLAM algorithms are evaluation with XTDrone.

Nowadays, there are some popular date sets such as EuRoC [20] and TUM Dataset [21], and different visual SLAM algorithms are evaluated on these data sets. However, to evaluate algorithms deeply, it is expected to supply a virtual real-time environment, in which parameters of camera, communication bandwidth, light, texture richness and action of camera can be adjusted and controlled by users. Therefore, we design a workflow to evaluate slam algorithms with XTDrone.

Fig. 3 shows the workflow of evaluation, where messages in red are key messages for visual SLAM evaluation. The evaluation includes localization accuracy and computational complexity, both of which are essential for UAVs. Pose ground truth is provided not only for the accuracy evaluation, but also for the UAV state estimator. The user controls the UAV to follow a desired trajectory via keyboard, and meanwhile the stereo camera on the UAV sends stereo image messages to the SLAM module, and the SLAM module outputs pose messages and record them as ROSBag or TUM/KITTI/EuRoC file. At the same time, system monitor records CPU and memory usage by the SLAM module. After flight simulation, the user can analysis the localization accuracy by EVO [9].

---

[8] https://www.yuque.com/xtdrone/manual_en/sensor_model

Two state-of-the-art and open source SLAM algorithm, ORB-SLAM2 [10] and VINS-Fusion [11], are chosen for the evaluation. VINS-Fusion is indeed an visual initial odometry, and to be fair, we use VINS-Fusion without IMU. For the simulation close to reality, the parameters of stereo camera are set equal to the Aptina MT9V034 global shutter stereo camera [12]. Fig. 4 shows the flight simulation and Fig. 5 shows the processes of ORB-SLAM2 and VINS-Fusion without IMU. The complete flight video can be seen at https://www.yuque.com/xtdrone/demo/vslam_evaluation

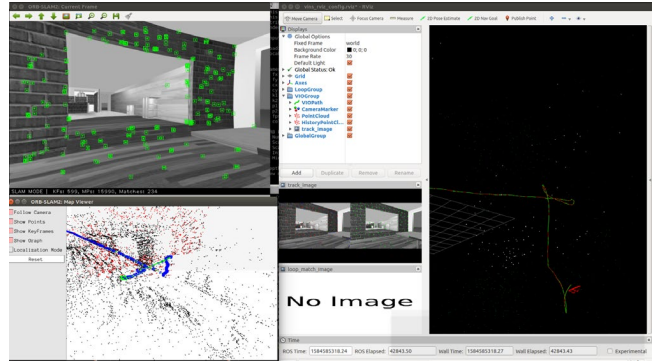

Figure 5.  ORB-SLAM2 (left) and VINS-Fusion without IMU (right)

Fig. 6 shows the trajectories generated by ORB-SLAM2 and VINS-Fusion without IMU, compared with the true flight trajectory. There are four large-scale yaw maneuverings, which is a great challenge to SLAM algorithms. Fig. 7 shows that the relative pose error (RPE) with regard to rotation angle. It can be seen that great errors happen when yaw maneuvering, and ORB-SLAM2 behaves better than VINS-Fusion without IMU.

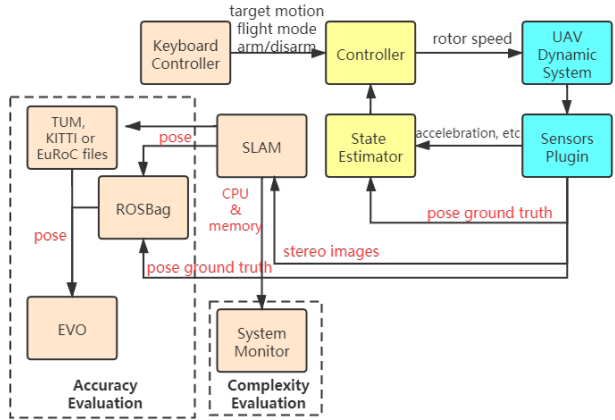

Figure 3.  Workflow of visual SLAM evaluation

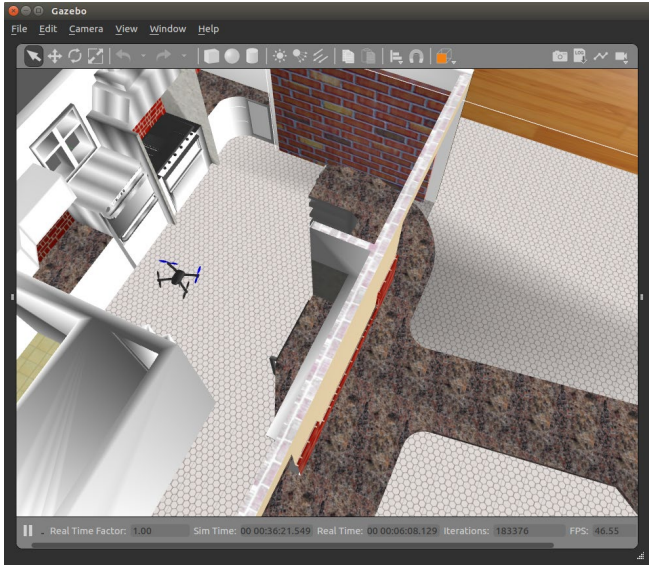

Figure 4.  Flight Simulation in Gazebo

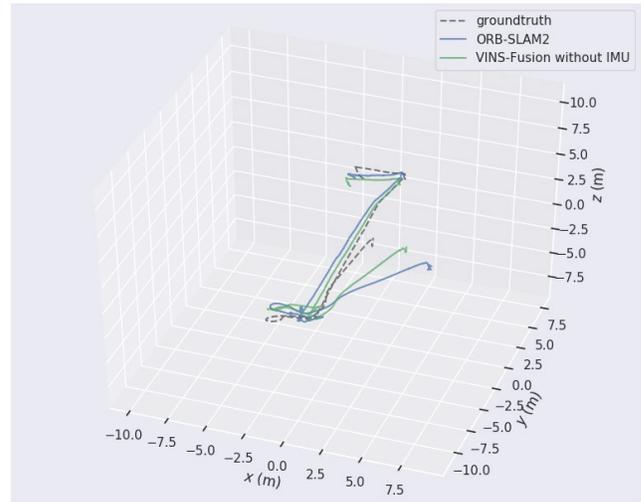

Figure 6.  Trajectories compared with the ground truth

---

[9] https://github.com/MichaelGrupp/evo
[10] https://github.com/raulmur/ORB_SLAM2
[11] https://github.com/HKUST-Aerial-Robotics/VINS-Fusion
[12] https://www.onsemi.com/pub/Collateral/MT9V034-D.PDF

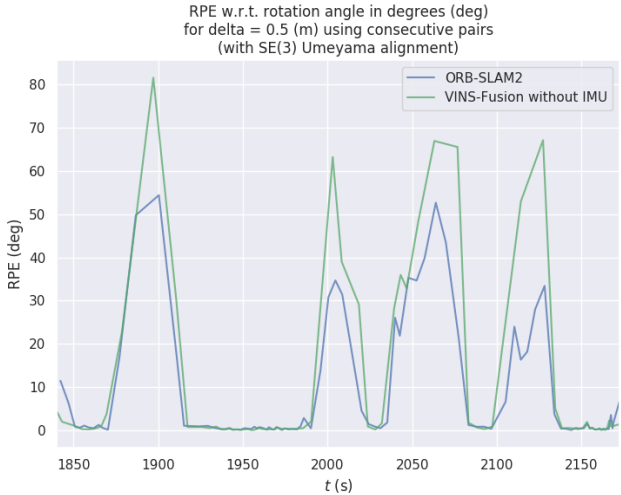

Figure 7. RPE w.r.t rotation angle

Relative pose error (RPE) and absolute pose error (APE) are two kind of error to evaluate accuracy of SLAM localization. RPE compares the pose increments between estimation and ground truth, which can give local accuracy, such as transition error and rotation error per meter. APE directly measures the difference between points of the true and the estimated trajectory, which can evaluate the global consistency of the estimated trajectory. For UAVs, RPE determines hovering precision and APE determines accuracy of mapping and motion planning.

Table III shows the comparison of accuracy between ORB-SLAM2 and VINS-Fusion without IMU. RPE is set to pose error per half a meter. As there is a little randomness in SLAM algorithms, five experiments whose condition is the same are carried out and RPE and APE are the average results. It can be seen that RPE of ORB-SLAM2 is much lower than that of VINS-Fusion without IMU but APE is a little higher.

TABLE III. COMPARISON OF ACCURACY

|  |  | *ORB-SLAM2* | *VINS-Fusion without IMU* |
|---|---|---|---|
| **RPE** | *Rotation (deg)* | 15.706 | 21.613 |
|  | *Translation (m)* | 0.525 | 0.714 |
| **APE** | *Rotation (deg)* | 129.229 | 126.168 |
|  | *Translation (m)* | 1.3196 | 1.0722 |

Table IV shows the comparison of computational complexity between ORB-SLAM2 and VINS-Fusion without IMU, where CPU usages evaluate time complexity and memory usages evaluate space complexity. The computation power of the simulation platform is DELL G3 with Intel i7-9750H CPU and 16GB DDR4 memory. Because SLAM algorithms can use multiple CPU kernels, the CPU usage can be larger than 100%. CPU and memory usage are average numbers among different experiments and different sampling time. It can be seen that VINS-Fusion without IMU has both less CPU usage and less memory usage, so it has less computational complexity.

TABLE IV. COMPARISON OF COMPUTATIONAL COMPLEXITY

|  | *ORB-SLAM2* | *VINS-Fusion without IMU* |
|---|---|---|
| *CPU* | 150.3% | 69.2% |
| *Memory* | 4.1% | 0.7% |

According to the evaluation, UAV developer may use ORB-SLAM2 if the UAV has a powerful chip but may use VINS-Fusion if the CPU and memory of chip are mediocre. SLAM is a kind of complex algorithms, and there are many parameters to adjust. This demo only shows the way to evaluate different SLAM algorithm and the parameters are almost set as default. For more careful evaluation, users are expected to be an expert at SLAM and adjust parameters in detail.

## V. MULTI-UAV FORMATION

Multi-UAV cooperation is a hot research field nowadays. A simulation platform is more urgently needed for the research of multiple UAVs, especially large-scale UAV swarm, than that of single UAV. In this section, we show the workflow of multi-UAV formation simulation with XTDrone.

Formation is a type of multi-UAV systems. A formation consists of cooperative interactions, and the relationship between UAVs is well-defined for objectives [22], such as forming a certain pattern for artistic performance. Leader-follower is a typical approach of flight formation control [23]. Leader UAV has its own trajectory and follower UAVs keep relative positions to the leader, so that a certain formation configuration is kept. There are two problems here, communication and control, and many researchers focus on how to devise communication scheme and control scheme. To be user-friendly, XTDrone has two python classes, one for the leader and another for the follower. Users can inherit these two classes and modify the communication scheme and control scheme. A typical communication topology realized in XTDrone simulation platform is shown as Fig. 8.

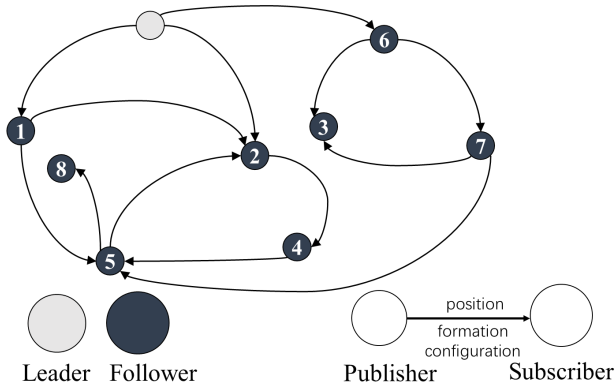

Figure 8.  Centralized formation scheme

A centralized formation scheme is the most mature scheme and is generally used in formation artist shows. It relies on the assumption that all the followers have the ability to communicate to the leader [24], thus the communication topology is simplified and the control can achieve global optimal [25]. Therefore, the multi-UAV formation demo uses the centralized scheme, shown as Fig. 9. The modules in dotted box are indeed motion planning modules shown in Fig. 1. Therefore, it's only needed to modify keyboard controller and motion planning modules, then the formation can be realized on the simulation platform.

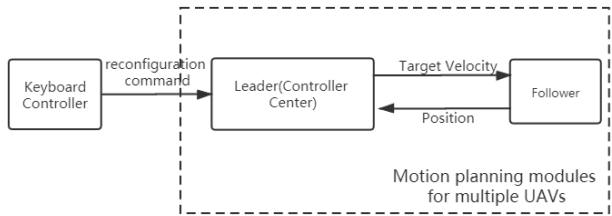

Figure 9.  Centralized formation scheme

The algorithm design and validation for UAV formation with XTDrone follows the workflow shown as Fig. 10. After designing communication and control scheme, users are expected to validate the formation on a simple 3D simulator. This simulator is especially designed with python and ROS, shown as Fig. 12. It subscribes target motion messages and publishes current motion messages, the topic names of which are the same as the standard XTDrone. The UAV dynamic model on the simple simulator is $2^{nd}$ order point mass dynamic model, which not only reduce computational complexity greatly, but also meet the demand of most formation control algorithm. After the formation algorithm works well in the simple 3D simulator, it will be validated in the XTDrone simulation platform, which has much more complex dynamic system and better graphical user interface, shown as Fig. 13. Through this workflow, Users can fast design and validate formation algorithm. Especially when the formation is large-scale, having hundreds even thousands of UAVs, simulation with XTDrone needs a powerful workstation otherwise the simulation speed will be very low.

Therefore, a preliminary verification in the simple 3D simulator is essential.

In this demo, one leader and eight followers form three configurations, cube, pyramid and triangle, shown as Fig. 11. We validate the centralized formation scheme in the simple 3D simulator, shown as Fig. 12, and in XTDrone, shown as Fig. 13. The complete flight video can be seen at https://www.yuque.com/xtdrone/demo/uav_formation

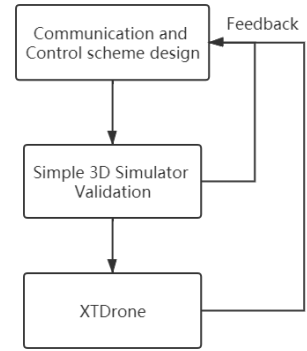

Figure 10.  Workflow of UAV formation design and validation

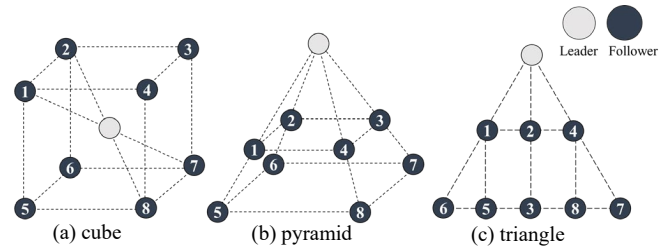

Figure 11.  Three formation configurations (grey cricle represents the leader and black circles represent the followers)

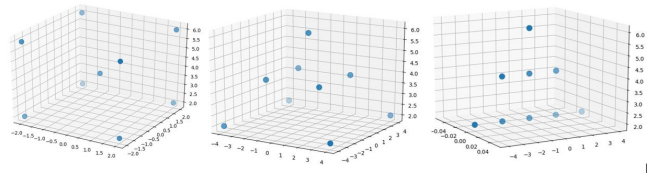

Figure 12.  Formation algorithm validation in the simple 3D simulator

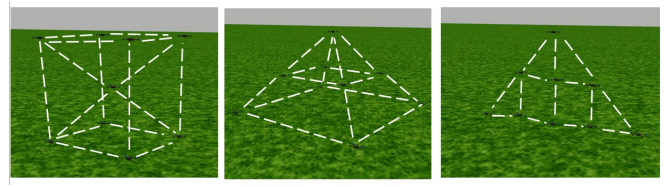

Figure 13.  Formation algorithm validation in XTDrone

QGroundControl [13] a powerful ground control station software and supports PX4 well. For the demo in this section, the whole UAV formation doesn't move and just does reconfiguration. For many missions, such as searching, the whole formation will move in a long distance. In that case, the trajectory of the leader can be conveniently planned by using QGroundControl and meanwhile the followers track the target motion sent from the leader as mentioned above.

---

[13] http://qgroundcontrol.com/

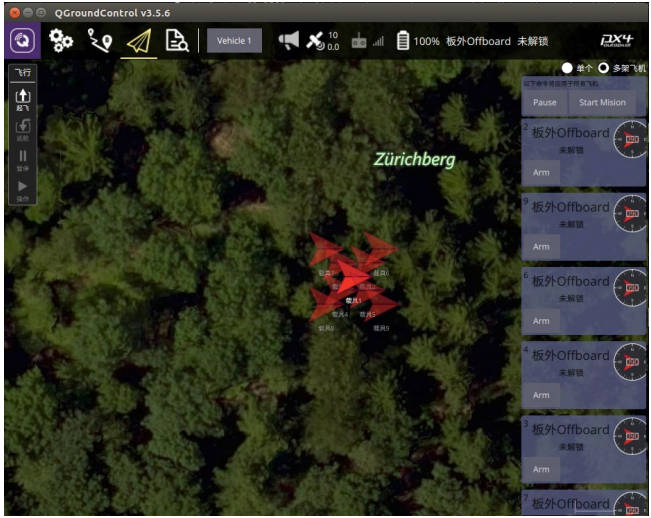

Figure 14. Multiple UAVs with QGroundControl

PID controllers are commonly used in formation controllers. Therefore, during the workflow shown in Fig. 10, PID parameters will be adjusted continually. For ROS, ROSBag is a useful tool to record logs, and for PX4, flight logs, containing almost every key message, are recorded as ULog [14] files. And there are some user-friendly software [15] to analysis ULog files. Fig. 15 shows the Y axis velocity tracking performance for Follower 1.

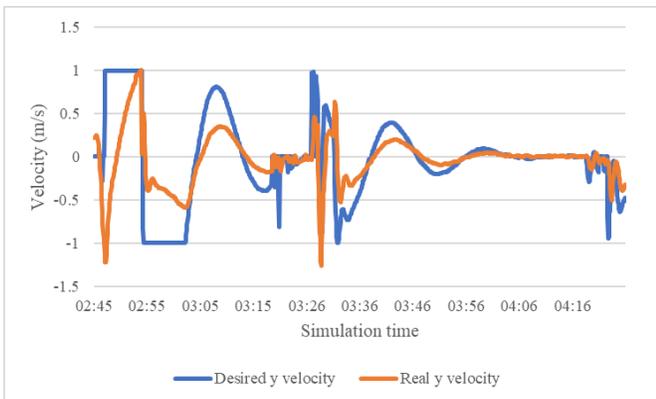

Figure 15. Y axis velocity tracking curves

---

[14] https://dev.px4.io/master/en/log/ulog_file_format.html
[15] https://docs.px4.io/master/en/log/flight_log_analysis.html

## VI. CONCLUSION AND FUTURE WORK

A customizable multi-rotor UAVs simulation platform based on ROS, Gazebo and PX4 is presented. The platform is modular so that users can test and deploy their algorithm conveniently. Two cases, evaluating different vision SLAM algorithm and realizing UAV formation, are shown to demonstrate the role of the platform and how the platform works.

The platform runs in lockstep, means that different numerical solvers maintain synchronized time, which makes it possible to run the simulation faster or slower than real time, and also to pause it in order to step through code. A powerful computer runs faster and not powerful one runs slower. Therefore, this feature makes computers with different performance possible to be used for simulation. Moreover, a simple 3D simulator is provided for multi-UAV simulation, thus speeding up the formation algorithm design.

The platform is still being improved, and more functions will be realized based on this platform in the future, such as controlling the UAV armed with a robot arm, multi-UAV cooperative construction and motion planning based on reinforcement learning. And we plan to hold a UAV challenge competition based on XTDrone.